\documentclass[10pt,twocolumn,letterpaper]{article}

\usepackage[]{cvpr}      

\usepackage{graphicx}
\usepackage{bbding}
\usepackage{amsmath}
\usepackage{amssymb}
\usepackage{booktabs}
\usepackage{float}
\usepackage{multirow}
\usepackage{bm}

\usepackage[pagebackref,breaklinks,colorlinks]{hyperref}
 
\usepackage[capitalize]{cleveref}
\crefname{section}{Sec.}{Secs.}
\Crefname{section}{Section}{Sections}
\Crefname{table}{Table}{Tables}
\crefname{table}{Tab.}{Tabs.}

\def\cvprPaperID{5260}
\def\confName{ICCV}
\def\confYear{2023}

\begin{document}

\title{Building3D: An Urban-Scale Dataset and Benchmarks for Learning Roof Structures from Point Clouds}

\author{Ruisheng Wang, Shangfeng Huang, Hongxin Yang\\
University of Calgary, AB, Canada\\
{\tt\small \{ruiswang, shangfeng.huang, hongxin.yang\}@ucalgary.ca}
}

\twocolumn[{
\renewcommand\twocolumn[1][]{#1}
\maketitle
\begin{center}
\captionsetup{type=figure}
  \begin{subfigure}{0.55\linewidth}
  \centering
    \includegraphics[width=1\linewidth]{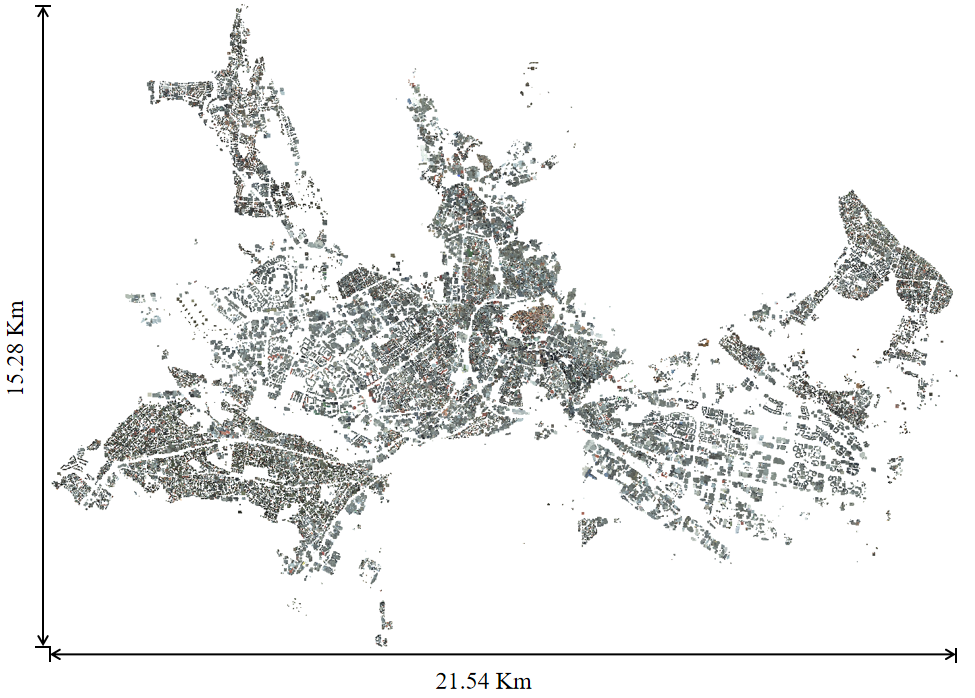}
    \caption{Aerial LiDAR point clouds of Tallinn}
    \label{fig_1:a}
  \end{subfigure}
  \hfill
  \begin{subfigure}{0.43\linewidth}
  \begin{subfigure}{1\linewidth}
  \begin{subfigure}{0.49\linewidth}
    \centering
    \includegraphics[width=1\linewidth]{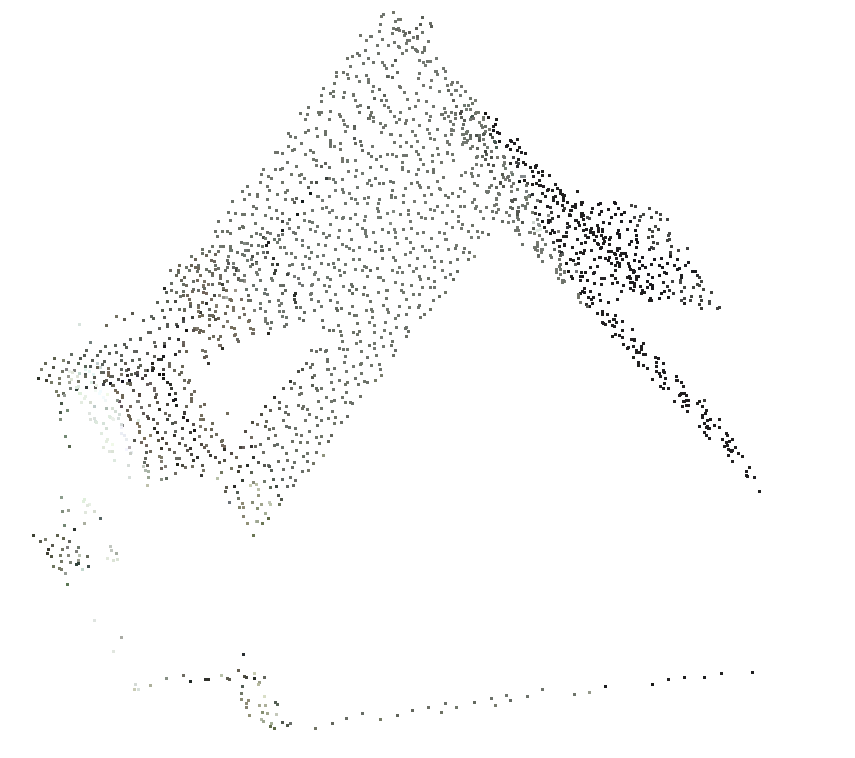}
    \caption{A building point cloud}
    \label{fig_1:b}
    \end{subfigure}
    \hfill
    \begin{subfigure}{0.48\linewidth}
    \centering
    \includegraphics[width=1\linewidth]{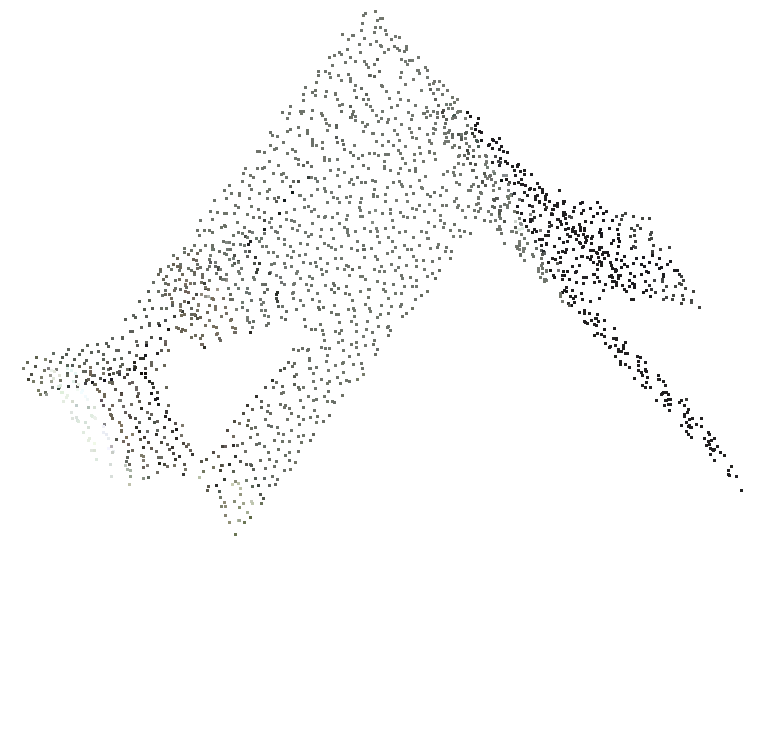}
    \caption{A roof point cloud}
    \label{fig_1:c}
    \end{subfigure}
  \end{subfigure}
  
  \begin{subfigure}{1\linewidth}
  \begin{subfigure}{0.48\linewidth}
  \centering
    \includegraphics[width=1\linewidth]{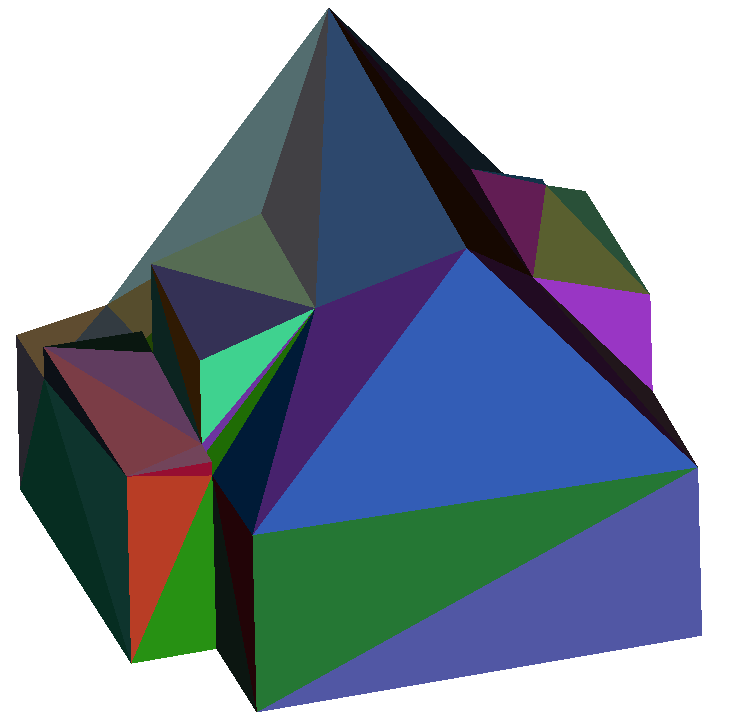}
    \caption{A mesh model}
    \label{fig_1:d}
  \end{subfigure}
  \hfill
  \begin{subfigure}{0.48\linewidth}
  \centering
    \includegraphics[width=1\linewidth]{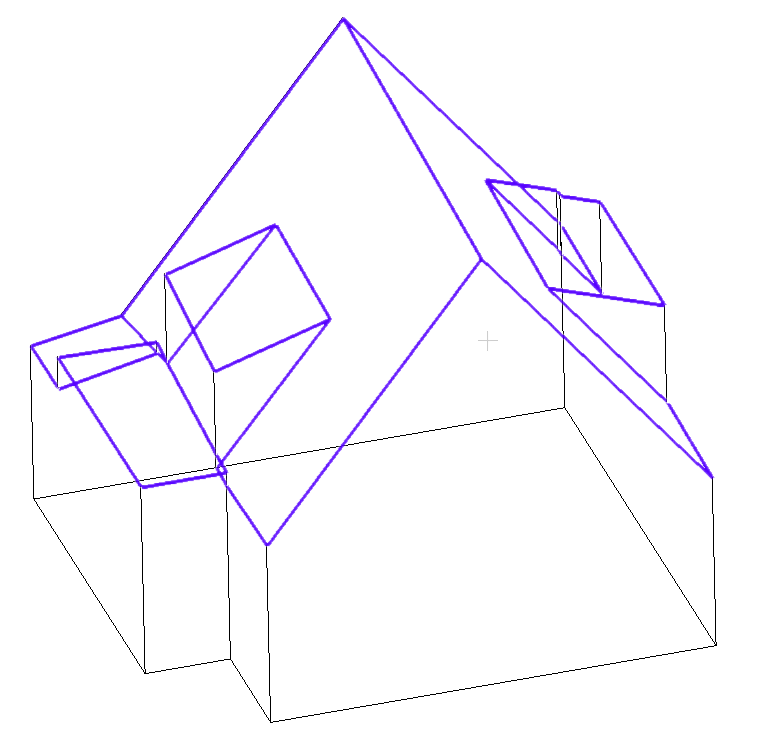}
    \caption{A wireframe model}
    \label{fig_1:e}
    \end{subfigure}
  \end{subfigure}
  
  \end{subfigure}
\caption{Illustration of Tallinn, one of the 16 cities in the Building3D dataset}
\label{fig_1}
\end{center}
}]
\begin{abstract}
Urban modeling from LiDAR point clouds is an important topic in computer vision, computer graphics, photogrammetry and remote sensing. 3D city models have found a wide range of applications in smart cities, autonomous navigation, urban planning and mapping etc. However, existing datasets for 3D modeling mainly focus on common objects such as furniture or cars. Lack of building datasets has become a major obstacle for applying deep learning technology to specific domains such as urban modeling.   
In this paper, we present an urban-scale dataset consisting of more than 160 thousands buildings along with corresponding point clouds, mesh and wireframe models, covering 16 cities in Estonia about 998 $Km^2$. 
We extensively evaluate performance of state-of-the-art algorithms including handcrafted and deep feature based methods. Experimental results indicate that Building3D has challenges of high intra-class variance, data imbalance and large-scale noises. The Building3D is the first and largest urban-scale building modeling benchmark, allowing a comparison of supervised and self-supervised learning methods. We believe that our Building3D will facilitate future research on urban modeling, aerial path planning, mesh simplification, and semantic/part segmentation etc.  

\end{abstract}


\section{Introduction}
\label{sec:intro}
Deep learning has achieved tremendous success on computer vision applications such as image classification and object detection\cite{NIPS2012_c399862d, brusilovsky:simonyan2014very, he2016residual, 47353}, semantic segmentation \cite{he2017mask,deeplabv3plus2018, ronneberger2015u} and human pose estimation \cite{8765346,wang2019deep}, to name a few. With the increased availability of 3D point clouds that found a wide range of applications in robotics, autonomous driving and urban modeling, a recent research focus has been shifted to deal with such massive 3D point clouds \cite{su15mvcnn,Maturana2015VoxNet,qi2017pointnet,qi2017pointnet++}. The majority of current work on 3D point cloud processing are focused on 3D shape classification \cite{su15mvcnn,thomas2019KPConv,Riegler2017OctNet, qi2017pointnet}, 3D object detection and tracking \cite{Maturana2015VoxNet,9652096,9156722}, 3D semantic and instance segmentation\cite{hu2019randla,thomas2019KPConv, qi2017pointnet++,Loic18}. Correspondingly, a large number of datasets including synthetic and real-world ones have been established to train and evaluate deep learning algorithms with respect to above mentioned applications \cite{chang2015shapenet,uy2019revisiting,6248074,9156412,semantic3D2017,isprs,hu2020towards, Chen_2022_BMVC}. The widely available datasets are prerequisite for rapid algorithm advancement in supervised learning based on neural networks. Supervised learning methods heavily rely on labeled data, which have been intensively studied. Due to the expensive cost of labeled data, self-supervised learning methods learning representations from unlabeled data, are receiving more and more attention. Started from self-supervised learning on 2D images, SimCLR \cite{pmlr-v119-chen20j} and CPC \cite{2018arXiv180703748V,DBLP:journals/corr/abs-1905-09272}
 have reached top performance on image classification benchmarks. The methods of self-supervised learning in 2D are being quickly adapted for 3D point clouds, such as the jigsaw puzzle pretext task \cite{sauder2019self}, estimating rotations \cite{kim1999robust}, contrastive learning \cite{xie2020pointcontrast}, point cloud completion \cite{wang2021unsupervised}. 
 

Current deep learning methods on urban modeling have been restricted to small datasets or synthetic ones. However, urban modeling is different from existing object modeling work where small objects are collected under a well-controlled lab environment. Specifically, urban modeling deal with large-scale LiDAR scans containing more noisy and incomplete point clouds that represent complex real-world scenes. To advance urban modeling research in computer vision, we introduce an urban-scale dataset for 3D roof modeling from point clouds collected from the air. The dataset covering 16 cities in Estonian consist of 875.39 millions of aerial LiDAR point clouds and 161.91 thousands of 3D building models in both mesh and wireframe formats. A mesh model is a 3D model that is made up of small discrete cells. The commonly used 2D cell shapes are the triangle and the quadrilateral, which the triangular mesh is the one referred in this paper. A wireframe model is a 3D model that the polygonal faces have been removed to retain only the outlines of its component polygons. It is the least complex representation, namely a skeletal description of a 3D object consisting of vector points connected by lines. Man-made objects such as buildings are mostly polyhedral which can be represented by corners, edges and/or planar surfaces\cite{Liu2019ICLR}. Therefore, 
wireframes are particularly suitable for representing polyhedral objects such as buildings or furniture. Besides benefit of efficient storage and transmit, wireframe models are easy to edit and manipulate in CAD software which can help create CAD models for various applications such as quality inspection, metrology, rendering and animation\cite{Liu2019ICLR}. 

We convert mesh models into wireframe models. There is no wireframe models provided in current building modeling datasets \cite{chang2015shapenet,wichmann2018roofn3d,huang2022city3d,lin2021capturing,hu2020towards}. To our knowledge, we are the first to provide both mesh and wireframe building models along with corresponding LiDAR point clouds at urban scale. Fig.\ref{fig_1:a} shows the aerial LiDAR point clouds of Tallinn, one out of 16 cities in Building3D dataset. It contains 361.95 million points and 47.05 thousands of buildings, covering an area around 195 km$^2$. Figure \ref{fig_1:b}, \ref{fig_1:c} \ref{fig_1:d} \ref{fig_1:e} shows building and roof point clouds, as well as corresponding mesh and wireframe models, respectively. 
Besides a urban-scale dataset, we also provide two new baselines of supervised and self-supervised learning, adopted and evaluated various supervised and self-supervised pipelines for 3D roof modeling. To our knowledge, we are the first to use and evaluate self-supervised learning method for 3D object reconstruction. Overall, our main contributions are in the following. 
\begin{itemize}
\item We present the first and largest urban-scale building modeling dataset consisting of aerial LiDAR point clouds, mesh and wireframe models. Besides urban modeling, the proposed dataset can be extended to support various downstream applications. The whole dataset is made available to the research community.
\item We evaluate representative deep and handcrafted feature based methods including mainstream self-supervised learning methods, and establish new baselines and evaluation metric for future research. The new baselines achieve state-of-the-art performance compared with deep learning based methods. To our knowledge, we are the first to propose and adopt self-supervised pre-training methods for 3D building reconstruction. 

\end{itemize}
 
\section{Related Work}
\label{sec:formatting}

\subsection{Related Datasets}

\begin{table*}[]
\centering
\setlength{\tabcolsep}{0.3mm}{
\begin{tabular}{@{}c|ccccccc@{}}
\toprule
Dataset   & Scene      &  Type      & Area($Km^2$)   & Diversity  & Point Clouds  & Mesh & Wireframe\\ \midrule
ShapeNetCore\cite{chang2015shapenet} & Object   & Synthetic   & --  & --  &  & \Checkmark  &  \\
KITTI\cite{geiger2012we}  & Street  & LiDAR  & --  &   39.2 Km street  &  \Checkmark &   &  \\
Simulated dataset\cite{li2022point2roof} & --  & Synthetic        & --               &  --  & \Checkmark & \Checkmark &    \\
RoofN3D by point2roof\cite{wichmann2018roofn3d} &  --       &     LiDAR &  --  & 1 city  &\Checkmark &\Checkmark &     \\
City3D\cite{huang2022city3d}  & Urban  & LiDAR  & --  & 3 scenes  &\Checkmark  &\Checkmark  &  \\
UrbanScene3D\cite{lin2021capturing} & Urban  & CAD\&MVS  & 136  &  \begin{tabular}[c]{@{}c@{}}6 cities and\\ 10 scenes \end{tabular}  &  & \Checkmark  &  \\
SensatUrban\cite{hu2020towards} & Urban & UAV Photogrammetry  & 7.64   & 3 cities &\Checkmark  & &\\
STPLS3D\cite{Chen_2022_BMVC} &Large scenes &Aerial Photogrammetry & 17 & 67 scenes & \Checkmark & \Checkmark &\\
DublinCity \cite{BMVC2019}  &Small Scenes  & LiDAR  &5.6 & 1 scene & \Checkmark & & \\
Building3D      & Urban       & LiDAR       &  998              &  16 cities  &  \Checkmark & \Checkmark  &\Checkmark \\
\bottomrule
\end{tabular}}
\caption{Comparison with existing 3D datasets }
 \label{tab:1}
\end{table*}

\begin{table*}[]
\centering
\setlength{\tabcolsep}{5mm}{
\begin{tabular}{@{}c|cccc@{}}
\toprule
\multirow{2}{*}{Method} & \multicolumn{4}{c}{Number}                \\
                        & Objects & Roof(avg.) & Corners (avg.) & Edges(avg.) \\ \midrule
Simulated dataset\cite{li2022point2roof}&  17.6 K       & 1409        & 8               & 10      \\
RoofN3D by point2roof\cite{wichmann2018roofn3d,li2022point2roof}& 0.5 K        & 1349      &  6              & 8      \\
Tallinn of Building3D      & 36.9 K       & 3292       &   16             & 18      \\ \bottomrule
\end{tabular}}
\caption{Quantitative comparison between Building3D and most relevant datasets}
 \label{tab:2}
\end{table*}

We only review pertinent 3D modeling datasets with a focus on point clouds based urban modeling datasets.  

\textbf{ShapeNet} \cite{chang2015shapenet} is currently one of the most popular 3D object datasets for part segmentation, point cloud completion and reconstruction. For 3D object reconstruction, ShapeNetCore, a subset of ShapeNet, covers 55 common object categories with about 51,300 unique 3D models but without corresponding point clouds.  
The methods \cite{park2019deepsdf,mescheder2019occupancy,peng2020convolutional} generate simulated point clouds as input, and use continuous signed distance field or 3D voxel grid with occupancy information for implicit reconstruction of 3D object surfaces.
\textbf{RoofN3D} \cite{wichmann2018roofn3d} is designed for 3D building reconstruction. 
We found that the quality of mesh models are poor and a large amount of shape and vertex are mismatched between point clouds and mesh models.
Based on RoofN3D dataset, Li et al., \cite{li2022point2roof} select 500 roof point clouds and manually create mesh models to construct a small real-world dataset on the top of a simulated dataset.
However, both the real-world and simulated datasets consist of very few categories of primitive shape, which doesn't exhibit considerable variability.  
\textbf{City3D}\cite{huang2022city3d}, a large-scale building reconstruction dataset, consists of about 20 thousands of building mesh models and aerial LiDAR point clouds. However, all mesh models are generated by the proposed method, which means that the quality of mesh models depends on the method's capability. A major limitation is that their framework uses only planar primitives. Therefore their method can't deal with curved surfaces or non-planar roofs which commonly exit in the real world. 
\textbf{UrbanScene3D} \cite{lin2021capturing} covers 10 synthetic and six real-world scenes. Specifically, 10 synthetic datasets consist of 13,352 textured mesh objects and corresponding aerial images for the sake of 3D reconstruction. The six real-world scenes provide 488 textured mesh objects and corresponding aerial images. Although point clouds are also provided, they are only allowed to be used for evaluating the quality of reconstruction results not for the training.
\begin{figure*}[h!]
\centering
\includegraphics[scale=0.46]{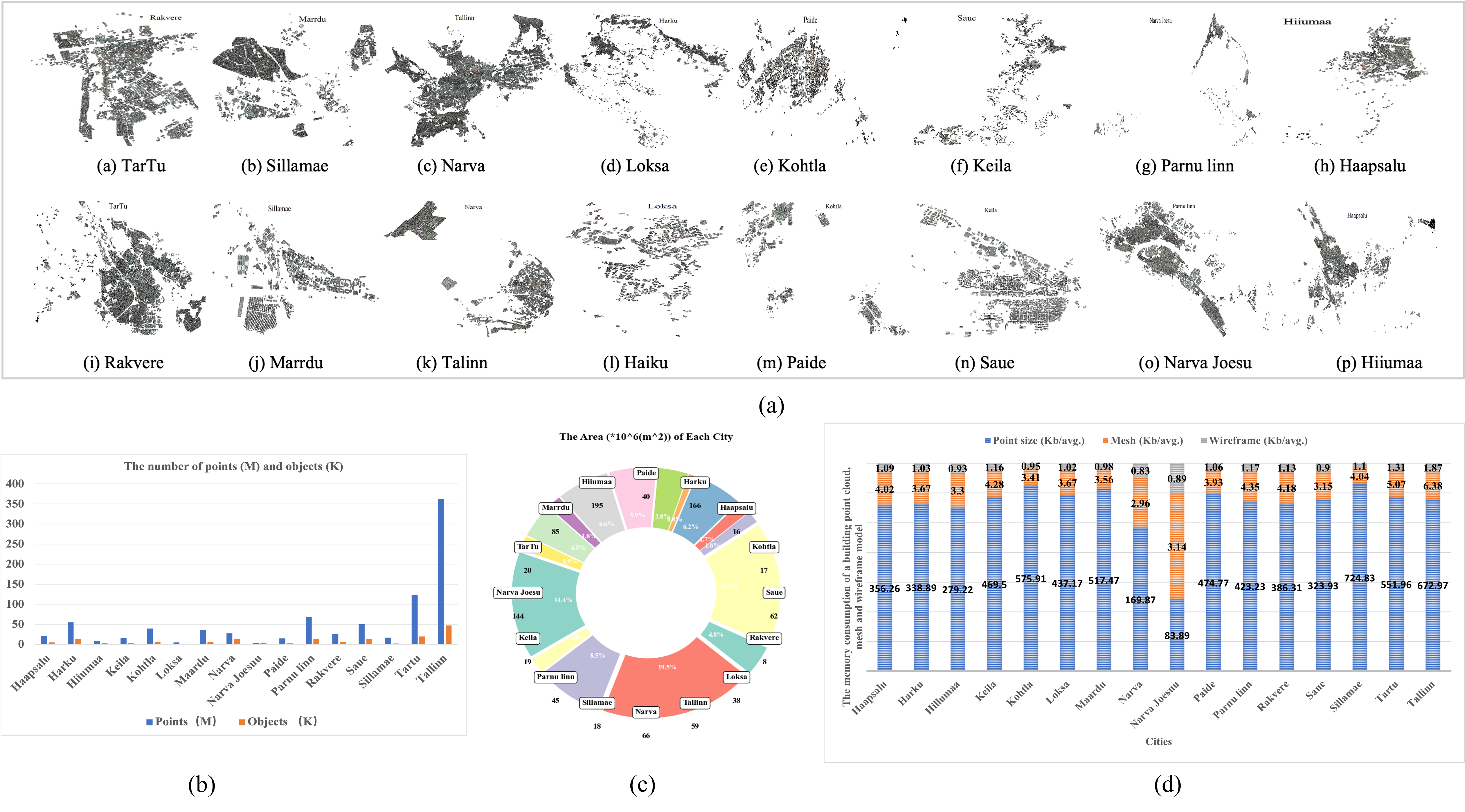}
\caption{Details of Building3D dataset. (a) Illustration of 16 cities in our dataset. (b) The number of points and objects in each city. (c) The area of each city. (d) The average memory consumption of a building point cloud, mesh and wireframe model in each city.}
\label{fig_5}
\end{figure*}
\subsection{Related Methods}
Building reconstruction from point clouds has been intensively studied in the past two decades \cite{poullis2009automatic, Nan2010, zhou20122, Vanegas2012, henn2013model, vosselman20013d, chen2017topologically, nan2017polyfit, PoullisPAMI2019, Minglei2016, li2022point2roof}. This problem remains unsolved due to complexity in roof structure and building style, sparsity, noise and possible missing data in point clouds. In general, traditional 3D roof reconstruction methods can be classified into three categories: data-driven \cite{poullis2009automatic, zhou20122}, model-driven \cite{henn2013model, vosselman20013d} and hybrid \cite{vosselman20013d}. Traditional methods normally require multiple steps of processing to generate 3D building models, and the errors introduced in each step will be accumulated. The quality of final building models largely rely on the quality resulted from previous steps. If an end-to-end deep learning method, which takes point clouds as input and outputs wireframe building models, is employed, the problem of accumulated errors can be eliminated to a large extent \cite{li2022point2roof}. However, there are very few deep learning methods to address this problem probably due to unavailability of a well-labeled and diversified dataset. 

The PC2WF \cite{Liu2019ICLR} is the first end-to-end deep learning approach to generate wireframe models from synthetic 3D point clouds. The problem of wireframe model generation is formulated as a problem of vertex and edge classification. Due to lack of a publicly available point cloud dataset, they constructed their own synthetic dataset. 
Along this line of work, the Point2Roof \cite{li2022point2roof}is the first trainable end-to-end deep neural network to generate 3D wireframe building models from LiDAR point clouds. Due to lack of a real-world dataset, they also used synthetic point clouds and models for the pre-training, and constructed a small amount of real world data  to fine tune the network. However, due to the limited training samples, their algorithm can only generate very limited types of building models. We believe that our urban-scale real-world dataset will help advance the state of art in the field of urban modeling.

\section{The Building3D Dataset}
We process the raw data provided by land board of republic of Estonia to generate Building3D dataset (Building 3D model data: Estonian Land Board 2022). The Building3D dataset covers 16 cities about 998 $Km^2$. Fig.\ref{fig_5} shows overall statistics of the proposed dataset, which contains about 160 thousands building point clouds with corresponding mesh and wireframe models. Fig.\ref{fig_5}b shows histograms of point clouds and objects (i.e. buildings) in each city. The order of magnitude of points and objects are indicated by symbols M (million) and K (thousand) respectively. Fig.\ref{fig_5}c shows the area of each city. Fig.\ref{fig_5}d shows the average memory consumption of a building point cloud, mesh and wireframe model in each city respectively. The average memory consumption for a building point cloud is between 83.89KB and 672.97KB, while the corresponding mesh and wireframe model is less than 6.5KB and 2KB. The ratio of average sizes among building point clouds, mesh and wireframe models is approximately 400:4:1. An straightforward visualization of Building3D is shown in \url{http://building3d.ucalgary.ca/}.



\subsection{Data Description and Annotation} 
\textbf{Building point clouds}
The raw LiDAR point clouds are collected by a high-precision RIEGL VQ-1560i scanner at altitude 2600 meters then stored in LAZ format. Each LAZ file covers 1 Km$^2$ and consists of  terrain, water, trees, bushes, buildings, bridges etc. The relative accuracy of point clouds is 20 mm. The density of point clouds is 30.314 points per square meter and the point-to-point distance
is 0.1816m. For 3D building reconstruction, we remove irrelevant point clouds and only retain building point clouds. To this end, point clouds and corresponding mesh models are projected onto XY plane. Then all irrelevant point clouds outside projected regions of mesh models are removed to obtain building point clouds. To generate fine point clouds for 3D building reconstruction as shown in Fig.\ref{fig_1:b}, only points whose shortest distances to corresponding mesh models are within a given threshold are retained. Each building point clouds are stored in XYZ format including XYZ coordinates, RGB color, near infrared information, intensity and reflectance. 

\textbf{Roof point clouds} 
It is inevitable that building facade are incomplete due to use of airborne scanners. In practice, some building point clouds have relatively complete facade information, while others may have barely information on facade. However, it is not an essential problem for 3D roof reconstruction which doesn't involve much of facade information. We can remove all the points representing facades. Basically, almost all facades are vertical to the XY plane. We calculate normals for each mesh face and remove all mesh faces whose normals are parallel to the XY plane to generate roof mesh models. A point is classified as roof if its distance to the roof mesh model is within a given threshold. The roof point clouds are shown in Fig.\ref{fig_1:c}.

\textbf{Mesh models}
Building mesh models are created from aerial LiDAR point clouds and building footprints by using the Terrasolid software with manual editing. These mesh models are typical LoD2 models including detailed and realistic representation of roofs. Compared with mesh models with dense triangular facets provided by ShapeNet\cite{chang2015shapenet} and UrbanScene3D\cite{lin2021capturing}, our mesh models can be considered as simplified mesh models with an average of 50 faces for each model. In the Tallinn dataset, approximately 40$\%$ of mesh models are less than 30 faces.  A mesh model is shown in Fig.\ref{fig_1:d} which different colors indicate different triangular meshes. 

\textbf{Wireframe models}
wireframe models are designed to formulate the problem of 3D reconstruction as a problem of point and edge classification. Compared with mesh models, wireframe models have clearer and more well-defined structures and require less disk space. To generate wireframe models, we calculate normals for each mesh. If the normals of two adjacent triangular meshes are approximately parallel, it indicates that they are co-planar and the shared edge will be removed. After all shared edges are removed, we obtain coarse wireframe models with redundant vertices. The coarse wireframe models also contain short edges that can be merged into long edges. Finally, the fine wireframe models are generated by removing all redundant vertices and merging short edges, then reviewed and adjusted by professional technicians. A wireframe roof model in the Building3D dataset is shown in Fig.\ref{fig_1:e}. Root Mean Square Error (RMSE) represents displacement between original point clouds and corresponding mesh and wireframe models and  the average RMSE is 0.065m.

\begin{figure*}[h!]
\centering
\includegraphics[scale=0.37]{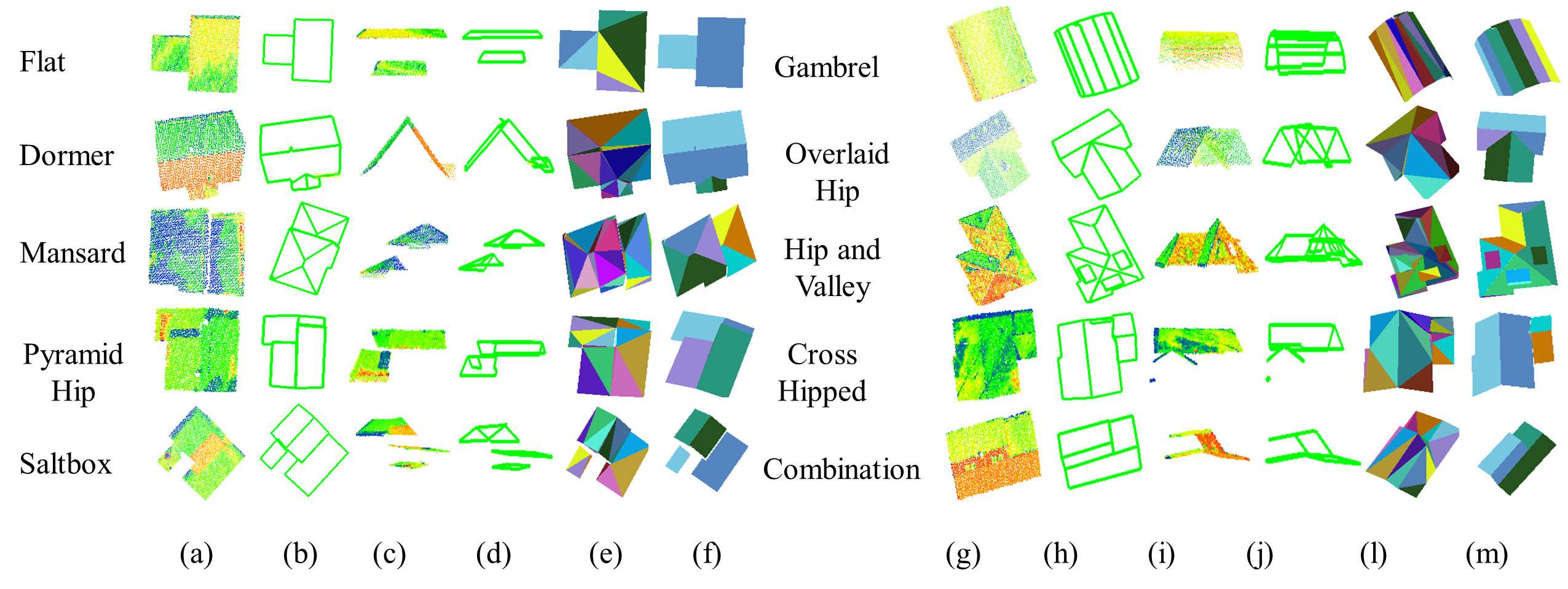}
\caption{Point cloud and wireframe models of 10 distinct roof types.}
\label{fig:4}
\end{figure*}

\subsection{Building Roof Types}
The Building3D dataset encompasses more than 60 distinct roof types, surpassing coverage of all comparable datasets as shown in Table \ref{tab:1}. Table \ref{tab:2} shows differences compared with most relevant datasets \cite{li2022point2roof,wichmann2018roofn3d}. The number of corners and edges shows that our roof models are more complex and have more categories. More points in roof point clouds indicate that our LiDAR point clouds contain more detailed information and intricate structure.  Fig.\ref{fig:4} illustrates some of the representative roof types, including Flat, Dormer, Mansard, Pyramid Hip, Saltbox, Glambrel, Overlaid Hip, Hip and Valley, Cross Hipped, Combination, among others. In each category of Fig.\ref{fig:4}, columns from left to right show roof point cloud and wireframe (top-down view), roof point cloud and wireframe (side view), corresponding roof mesh and wireframe model, respectively. The wireframe models contain less facets than mesh models.  

\section{Benchmarks}

\subsection{Evaluation metrics}  
We use several metrics for evaluation, average corner offset (ACO), precision, recall and F1 score, 3D mesh IoU, and root mean square error (RMSE). ACO is the average offset between predicted corners and ground-truth corners. 
Corner precision (CP), edge precision (EP), corner recall (CR), and edge recall (ER) are calculated through confusion matrix to evaluate the accuracy of corner and edge classification. 
3D Mesh IoU is a metric for evaluation of the fit between generated mesh models and ground truth mesh models. Existing 3D IoU methods use minimum bounding boxes of mesh models which can't represent accuracy of generated mesh models. We develop a numerical solution to use mesh models for 3D IoU to represent fitting errors.
RMS distance is a metric for evaluation of the fit between input roof point clouds and generated mesh models.

 \subsection{Training}
We carefully selected about 37k data samples consisting of roof point clouds and wireframe models from Tallinn. We use around 33.3k ($90\%$) data for the training and the remaining 3.7k ($10\%$) for the testing. The number of each roof point cloud is either upsampled or downsampled to 3072 as input, where the input $P\in R^{3072\times 7} $ contains  XYZ coordinates, RGB color and near infrared spectrum values. All models are trained in a RTX A6000 GPU with 48GB. In addition, we also select approximately 5,600 for training and 550 for testing as entry-level Building3D dataset, which contains approximately 10 corners and 12 edges per building on average.

\subsection{Representative Baselines}
To our knowledge, there are only two deep learning methods \cite{Liu2019ICLR, li2022point2roof} available for wireframe reconstruction from point clouds. This is probably due to limited availability of real-world datasets. The PC2WF \cite{Liu2019ICLR} is designed to generate furniture models which requires very dense point clouds as input. It doesn't work with point clouds of lower density and fails on the Building3D dataset. We evaluate performance of Point2Roof and also propose a new supervised method. Besides evaluation of four representative self-supervisions and three supervisions, we also propose a new self-supervised feature extraction module. In total, we select ten representative methods as solid baselines to benchmark our Building3D dataset.  

$\bullet$ PointNet\cite{qi2017pointnet} and PointNet++ \cite{qi2017pointnet++}: The most popular methods to extract 3D point cloud features.  

$\bullet$ RandLA-Net\cite{hu2019randla}: A method that can extract large-scale point cloud features by an effective local feature aggregation module.

$\bullet$ Point2Roof \cite{li2022point2roof}: It employs the PointNet++ \cite{qi2017pointnet++} as feature extractor to generate pointwise point features for roof corner prediction and edge classification. 
 
$\bullet$ Our Supervised: We use Point-Transformer \cite{zhao2021point} as feature extractor to generate pointwise point features for roof corner prediction and feature generation. Based on graph neural networks\cite{zhou2020graph},our method makes full use of node (corners) location and feature information to generate edges (refer to our supplementary material). 
 
$\bullet$ Point-BERT \cite{yu2022point}: A prior baseline that introduces a new pre-training approach using a masked point modeling pretext task for 3D point cloud Transformer.
 
$\bullet$ Point-MAE \cite{pang2022masked}: A solid baseline that introduces a self-supervised learning method for point clouds using masked auto-encoders.
 
$\bullet$ Point-M2AE \cite{zhang2022point}: It introduces a hierarchical pre-training method for point clouds using multi-scale masked auto-encoders.
 
$\bullet$ 3D-OAE \cite{zhou2022self}: It uses occlusions in the point clouds to train an auto-encoder and outperforms existing self-supervised methods for point cloud classification and segmentation tasks.

$\bullet$ {Our Self-supervised}: Based on Point-MAE, we designed a new linear self-attention mechanism to increase computational efficiency (refer to our supplementary material). 

\begin{table*}[]
\centering
\setlength{\tabcolsep}{4mm}{
\begin{tabular}{@{}cc|c|ccc|ccc@{}}
\toprule
\multicolumn{2}{c|}{Methods}                  & ACO & CP & CR & F1 & EP & ER & F1 \\ \midrule
\multirow{5}{*}{Self-supervised} & Point-BERT\cite{yu2022point} & 0.25 & 0.88 & 0.69 & 0.77 & 0.90 & 0.29  & 0.44   \\
                                 & Point-MAE \cite{pang2022masked} & 0.27 & 0.85  & 0.69 & 0.76 & 0.86  & 0.22 & 0.35   \\
                                 & Point-M2AE \cite{zhang2022point}& 0.26  & 0.88 & 0.69 & 0.77  & 0.90  & 0.31 & 0.46  \\
                                 & 3D-OAE \cite{zhou2022self} & 0.27 &  0.86 & 0.68 & 0.76 & 0.79 & 0.32 & 0.46   \\
 & \begin{tabular}[c]{@{}c@{}}Our self-supervised\\ (Based on Linear Transformer)\end{tabular} & 0.26  & \textbf{0.87} &\textbf{0.69} &\textbf{0.77}  & 0.87 & \textbf{0.37}  & \textbf{0.52} \\ \midrule
\multirow{4}{*}{Supervised}      & PointNet\cite{qi2017pointnet}   & 0.36  & 0.71 & 0.50  & 0.59  &0.81  & 0.26  &0.39  \\
                                 & PointNet++ \cite{qi2017pointnet++}& 0.34     &0.79    & 0.52 & 0.63 & 0.84   &0.33   &0.47 \\
                                & RandLA-Net \cite{hu2019randla}  & 0.35     & 0.70  & 0.60 &0.65  &0.67  & 0.16   &  0.25  \\
                                &DGCNN \cite{phan2018dgcnn}   & 0.32 & 0.73 &0.58 &0.65 & 0.81 &0.30 &0.44\\ 
                                &PAConv \cite{xu2021paconv} & 0.33 & 0.75 &0.57 &0.65 & 0.85 &0.31 &0.45\\ 
                                &Stratified Transformer \cite{lai2022stratified} & 0.38 & 0.72 &0.51 &0.62 & 0.75 &0.22 &0.34\\ 
                                &FG-Net \cite{Liu2022FGNetAF}    &0.32  &0.77 &0.64  &0.70  &0.84 &0.38  &0.52 \\
                                 & Point2RooF\cite{li2022point2roof} &0.30     &0.66    &0.48  &0.56   &0.71    &0.26  &0.38  \\
 & \begin{tabular}[c]{@{}c@{}}Our supervised\\ (Based on Point Transformer)\end{tabular}       &\textbf{0.26}  & \textbf{0.89} & 0.66 &\textbf{0.76} &\textbf{0.91} &\textbf{0.46} & \textbf{0.61}\\ \bottomrule
\end{tabular}}
\caption{Quantitative results on the entry-level Building3D}
\label{table:3}
\end{table*}

\begin{table*}[]
\captionsetup{skip=2pt}
\centering
\setlength{\tabcolsep}{4mm}{
\begin{tabular}{@{}cc|c|ccc|ccc@{}}
\toprule
\multicolumn{2}{c|}{Methods}                  & ACO & CP & CR & F1 & EP & ER & F1 \\ \midrule
\multirow{5}{*}{Self-supervised} & Point-MAE (1\%)  &  0.49 & 0.21 & 0.07 & 0.11 & 0.00 &0.00  &0.00 \\
                                 & Point-MAE (10\%) & 0.38 & 0.69  & 0.42 & 0.52 & 0.00  & 0.00  & 0.00  \\
                                 & Point-MAE (20\%) & 0.36 & 0.71  & 0.44 & 0.54 & 0.47  & 0.09  & 0.15  \\
                                 & Point-MAE (40\%) & 0.34 & 0.73  & 0.46 & 0.56 & 0.49  & 0.10  & 0.17  \\
                                 & Point-MAE (50\%) & 0.33 & 0.75  & 0.47 & 0.58 & 0.52  & 0.12  & 0.20  \\ \midrule
\multirow{5}{*}{Self-supervised} & Point-M2AE (10\%)  & 0.38 & 0.69 & 0.52 & 0.59 & 0.42 & 0.02   & 0.04\\
& Point-M2AE (20\%)  & 0.35 & 0.73  & 0.55 & 0.63 & 0.39  & 0.05 & 0.09\\
& Point-M2AE (40\%) & 0.32 & 0.77  & 0.57  & 0.66 & 0.42  & 0.08 & 0.13\\
& Point-M2AE (50\%) & 0.32 &  0.79 & 0.58 & 0.67 & 0.50 & 0.07  & 0.12 \\ \midrule
\multirow{5}{*}{Self-supervised} & Our self-supervised (1\%) & 0.57  & 0.34 & 0.04 & 0.07  &0.13  & 0.00 &0.00  \\
& Our self-supervised (10\%)  & 0.39 & 0.71 & 0.46 & 0.56 & 0.60 & 0.01 & 0.02\\
& Our self-supervised (20\%)  & 0.37 & 0.76  & 0.49 & 0.64 & 0.78  & 0.12 & 0.21\\
& Our self-supervised (40\%)  & 0.32 & 0.79  & 0.51 & 0.62 & 0.82  & 0.14 & 0.24 \\
& Our self-supervised (50\%)& 0.30 & 0.84  & 0.53  &0.65 & 0.85  & 0.15 &0.26\\
& Our self-supervised (80\%) & \textbf{0.28} &  \textbf{0.87} & \textbf{0.55} & \textbf{0.67} & \textbf{0.89}& \textbf{0.16}  & \textbf{0.27}
 \\ \midrule
\multirow{4}{*}{Supervised} & Point2RooF\cite{li2022point2roof} & 0.39 & 0.65 & 0.30 &0.41 & 0.66 & 0.08  & 0.14 \\
 & \begin{tabular}[c]{@{}c@{}}Our supervised\\ (Based on Point Transformer)\end{tabular}       &  \textbf{0.29}    &\textbf{0.90}     & \textbf{0.53}    &\textbf{0.66}  &\textbf{0.88}    & \textbf{0.23}  &\textbf{0.36}\\ \bottomrule
\end{tabular}}
\caption{Quantitative results on the Tallinn Building3D}
\label{table:4}
\end{table*}

\begin{table*}[]
\captionsetup{skip=2pt}
\centering
\renewcommand{\arraystretch}{1.2}
\setlength{\tabcolsep}{1.2mm}{
\begin{tabular}{@{}ccllcllcllcll@{}}
\toprule
\multirow{2}{*}{Method} &
  \multicolumn{3}{c}{Building A} &
  \multicolumn{3}{c}{Building B} &
  \multicolumn{3}{c}{Building C} &
  \multicolumn{3}{c}{Building D} \\ \cmidrule(l){2-13} 
 &
  \multicolumn{3}{c}{RMSE $\downarrow$ / IoU $\uparrow$  / Faces$\downarrow$} &
  \multicolumn{3}{c}{RMSE $\downarrow$ / IoU $\uparrow$  / Faces$\downarrow$} &
  \multicolumn{3}{c}{RMSE $\downarrow$ / IoU $\uparrow$  / Faces$\downarrow$} &
  \multicolumn{3}{c}{RMSE $\downarrow$ / IoU $\uparrow$  / Faces$\downarrow$} \\ \midrule
\begin{tabular}[c]{@{}c@{}}2.5D Dual\\ Contouring\cite{zhou20122}\end{tabular}    & \multicolumn{3}{c}{0.091\hskip 0.1cm/ \hskip 0.1cm 0.92\hskip 0.1cm /\hskip 0.1cm 268} & \multicolumn{3}{c}{ 0.134 \hskip 0.1cm / \hskip 0.1cm \textbf{0.73} \hskip 0.1cm/\hskip 0.1cm 100} & \multicolumn{3}{c}{0.091\hskip 0.1cm/\hskip 0.1cm0.92\hskip 0.1cm/ \hskip 0.1cm96} & \multicolumn{3}{c}{\textbf{0.094}\hskip 0.1cm/\hskip 0.1cm\textbf{0.95}\hskip 0.1cm/\hskip 0.1cm 283 } \\
PolyFit \cite{nan2017polyfit}                  & \multicolumn{3}{c}{\textbf{0.080}\hskip 0.1cm/\hskip 0.1cm \textbf{0.97} \hskip 0.1cm/\hskip 0.1cm 31  } & \multicolumn{3}{c}{\textbf{0.111}\hskip 0.1cm/ \hskip 0.1cm0.36 \hskip 0.1cm/ 158} & \multicolumn{3}{c}{\textbf{0.055}\hskip 0.1cm/ \hskip 0.1cm\textbf{0.97}\hskip 0.1cm / \hskip 0.1cm73} & \multicolumn{3}{c}{0.122\hskip 0.1cm/\hskip 0.1cm0.63\hskip 0.1cm/\hskip 0.1cm35} \\
\begin{tabular}[c]{@{}c@{}}Topology Aware\\ Modeling\cite{chen2017topologically}\end{tabular} & \multicolumn{3}{c}{0.114\hskip 0.1cm/\hskip 0.2cm -\hskip 0.3cm /\hskip 0.1cm3022 } & \multicolumn{3}{c}{0.346\hskip 0.1cm /\hskip 0.2cm 0.56\hskip 0.1cm /\hskip 0.1cm320} & \multicolumn{3}{c}{0.40\hskip 0.1cm /\hskip 0.1cm 0.73\hskip 0.15cm /\hskip 0.1cm355 } & \multicolumn{3}{c}{0.22\hskip 0.1cm /\hskip 0.2cm - \hskip 0.2cm/\hskip 0.1cm1373} \\
Groud Truth             & \multicolumn{3}{c}{0.086\hskip 0.1cm /\hskip 0.2cm - \hskip 0.2cm/ 80} & \multicolumn{3}{c}{0.171\hskip 0.1cm /\hskip 0.2cm - \hskip 0.2cm /\hskip 0.1cm54} & \multicolumn{3}{c}{0.052\hskip 0.1cm /\hskip 0.2cm - \hskip 0.2cm/\hskip 0.1cm68} & \multicolumn{3}{c}{0.063\hskip 0.1cm / \hskip 0.2cm - \hskip 0.2cm/\hskip 0.1cm54} \\
Ours                    & \multicolumn{3}{c}{0.490\hskip 0.1cm/\hskip 0.1cm 0.73\hskip 0.1cm /\hskip 0.1cm\textbf{19}} & \multicolumn{3}{c}{0.263 \hskip 0.1cm/\hskip 0.1cm 0.67\hskip 0.1cm / \hskip 0.1cm\textbf{13}} & \multicolumn{3}{c}{0.133\hskip 0.1cm /\hskip 0.1cm0.93\hskip 0.1cm/ \hskip 0.1cm\textbf{17}} & \multicolumn{3}{c}{0.096\hskip 0.1cm / \hskip 0.1cm0.92\hskip 0.1cm /\hskip 0.1cm\textbf{27} } \\ \bottomrule
\end{tabular}}
\caption{Quantitative results with traditional methods}
\label{table:5}
\end{table*}
\begin{figure*}
\captionsetup{skip=2pt}
    \centering
    \includegraphics[width=1\textwidth]{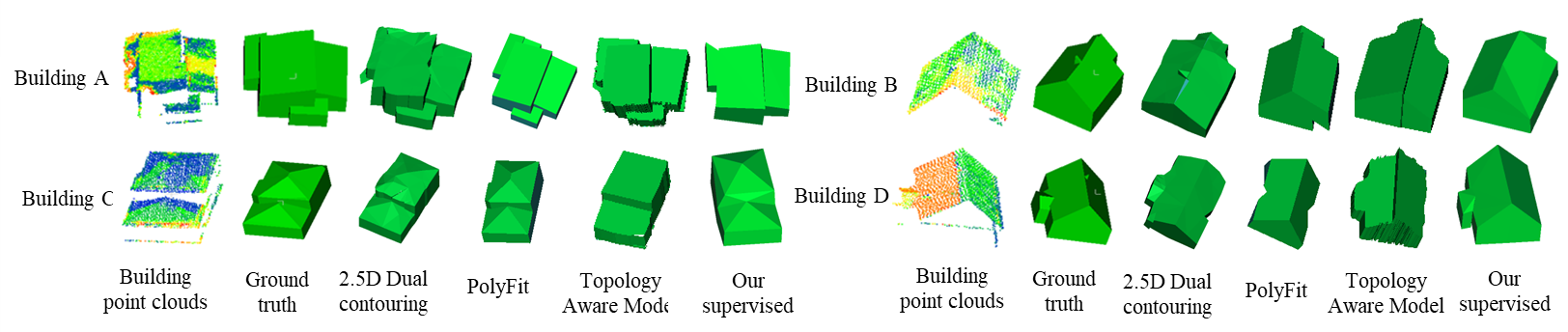}
    \caption{The visualization results from three traditional and our methods}
    \label{fig:4}
\end{figure*}




\subsection{Benchmark Results}
\textbf{Baselines On Deep Learning Methods} Table \ref{table:3} shows quantitative results of supervised and self-supervised methods with different feature extractors on the entry-level Building3D dataset. Although these methods show good performance in CP and EP metrics, they tend to perform poorly in CR and ER metrics, particularly in ER metric. This indicates that these methods struggle to detect all edges in the wireframe models, resulting in missing edges. In general, our supervised method performs the best in terms of F1 score in the supervised learning category. Our self-supervised performs the best in corner F1 score. Table \ref{table:4} shows that our supervised method outperforms the only existing deep-learning method \cite{li2022point2roof} in roof reconstruction by a large margin on the Tallinn Building3D. This is probably because Point2Roof was designed without sufficient real-world data for training, which limits its applicability to real-world scenarios beyond simulated data. As shown in Table \ref{table:3} and Table \ref{table:4}, all methods don't perform well in the ER metric, probably because of data incompleteness such as missing corner point clouds and insufficient edge feature extraction. This is also one of the major challenges in the Building3D dataset. For the self-supervised baseline, we use the pre-trained self-supervised module to replace the supervised feature extractor\cite{zhao2021point}. Then use partial labeled data at a reduced ratio, such as 1\%, 10\%, 20\%, 50\% and 80\% to fine tune the network to test its performance. Table \ref{table:4} indicate that the performance of self-supervised method is increased when using more labeled data. Our method achieves generally better performance at the same reduced data ratios compare to the Point-MAE\cite{pang2022masked}. More experiments are provided in our supplementary material. 


{\textbf{Modeling Results on Handcrafted Features} We evaluate three representative methods, 2.5D Dual Contouring \cite{zhou20102}, PolyFit \cite{nan2017polyfit}, and Topology Aware Modeling \cite{chen2017topologically}. 
These traditional methods typically require intensive parameter tuning from experts who well understand the algorithms, and errors in each step are accumulated which affect the final reconstruction. Specifically, the PolyFit requires complete facade point cloud information for the reconstruction process.  The 2.5D Dual Contouring and Topology Aware Modeling require denser point clouds as input.
On the contrary, our deep learning method is fed into open point clouds and can be used by non-professionals without any parameter tuning. 
Table \ref{table:5} shows quantitative results compared with the traditional methods. 
Although our method doesn't surpass traditional methods in terms of RMSE and IoU, the generated models have fewer faces, resulting smoother models as shown in Fig.\ref{fig:4}. This is because the ground truth wireframe in the Building3D dataset serves as guide to generate smoother mesh models.  

\section{Multi-purpose of Building3D Dataset}

\textbf{Building Semantic \& Part Segmentation} 
Our building point clouds, roof point clouds and facade point clouds have been assigned unique labels when building the roof reconstruction dataset. Therefore, Building3D can be extended for evaluating semantic segmentation and part segmentation of buildings. Large-scale real-world point cloud processing are very challenging in terms of semantic segmentation and part segmentation tasks etc. 
 
\textbf{Mesh Simplification} 
 Building3D dataset provides a large collection of building point clouds that can be triangulated into continuous and intricate mesh models. These triangulated mesh models can be used as original input for mesh simplification. Moreover, the quality of mesh simplification can be evaluated by calculating 3D mesh IoU between ground truth mesh models provided by Building3D dataset and simplified mesh models. 

\textbf{Footprint Detection} 
In our dataset, building models are generated by using 2D building footprints and aerial LiDAR point clouds. The Building3D dataset provides 2D building footprints as ground truth labels, which can be used to train and evaluate building footprint extraction methods from aerial LiDAR point clouds. Building footprint extraction from aerial LiDAR point clouds or images is important for producing maps which can be used in various applications such as urban planning and navigation.

\textbf{Aerial Path Planning}  
 Aerial path planning is a process of planning a path from starting point to target point with minimum energy consumption and collision avoidance. It considers factors such as potential threats of collision and path length. Building3D dataset provides large-scale 3D urban models that can be used as obstacle factors to train drones for finding optimal paths.

\section{Conclusions and Future Work}
In this paper, we present an urban-scale dataset for building roof modeling from aerial LiDAR point clouds. It consists of more than 160 thousands buildings, covering about 998 $Km^2$ of urban landscape. Besides mesh models and real-world LiDAR point clouds, it is the first time to release wireframe models which transforms 3D building reconstruction into a classification problem. We also provide two new baselines, a supervised and a self-supervised learning method, allowing a fair comparison between two learning modalities. The evaluation results indicate that our dataset is challenging and creates new opportunities for urban modeling research. We believe that this work will help advance future research on several fundamental problems as well as common object modeling such as mesh simplification and remeshing. In the future work, we aim to add detailed building facade models to enable LoD3 modeling for photorealistic building model generation, and associate address data to each building for holistic 3D scene understanding. 

{\small
\bibliographystyle{ieee}
\bibliography{egbib}

\begin{thebibliography}{10}\itemsep=-1pt

\bibitem{NIPS2012_c399862d}
A.~Krizhevsky, I.~Sutskever, and G.~E. Hinton.
\newblock Imagenet classification with deep convolutional neural networks.
\newblock In F.~Pereira, C.~Burges, L.~Bottou, and K.~Weinberger, editors, {\em
  Advances in Neural Information Processing Systems}, volume~25. Curran
  Associates, Inc., 2012.

\bibitem{brusilovsky:simonyan2014very}
K.~Simonyan and A.~Zisserman.
\newblock {Very deep convolutional networks for large-scale image recognition}.
\newblock {\em arXiv preprint arXiv:1409.1556}, 2014.

\bibitem{he2016residual}
K.~He, X.~Zhang, S.~Ren, and J.~Sun.
\newblock {Deep Residual Learning for Image Recognition}.
\newblock In {\em Proceedings of 2016 IEEE Conference on Computer Vision and
  Pattern Recognition}, CVPR '16, pages 770--778. IEEE, June 2016.

\bibitem{47353}
C.~Liu, B.~Zoph, M.~Neumann, J.~Shlens, W.~Hua, J.~Li, F.-F. Li, A.~Yuille,
  J.~Huang, and K.~Murphy.
\newblock Progressive neural architecture search.
\newblock 2018.

\bibitem{he2017mask}
K.~He, G.~Gkioxari, P.~Doll{\'a}r, and R.~Girshick.
\newblock Mask r-cnn.
\newblock In {\em Proceedings of the IEEE international conference on computer
  vision}, pages 2961--2969, 2017.

\bibitem{deeplabv3plus2018}
L.-C. Chen, Y.~Zhu, G.~Papandreou, F.~Schroff, and H.~Adam.
\newblock Encoder-decoder with atrous separable convolution for semantic image
  segmentation.
\newblock In {\em ECCV}, 2018.

\bibitem{ronneberger2015u}
O.~Ronneberger, P.~Fischer, and T.~Brox.
\newblock U-net: Convolutional networks for biomedical image segmentation.
\newblock In {\em International Conference on Medical image computing and
  computer-assisted intervention}, pages 234--241. Springer, 2015.

\bibitem{8765346}
Z.~{Cao}, G.~{Hidalgo Martinez}, T.~{Simon}, S.~{Wei}, and Y.~A. {Sheikh}.
\newblock Openpose: Realtime multi-person 2d pose estimation using part
  affinity fields.
\newblock {\em IEEE Transactions on Pattern Analysis and Machine Intelligence},
  2019.

\bibitem{wang2019deep}
J.~Wang, K.~Sun, T.~Cheng, B.~Jiang, C.~Deng, Y.~Zhao, D.~Liu, Y.~Mu, M.~Tan,
  X.~Wang, W.~Liu, and B.~Xiao.
\newblock Deep high-resolution representation learning for visual recognition.
\newblock {\em TPAMI}, 2019.

\bibitem{su15mvcnn}
H.~Su, S.~Maji, E.~Kalogerakis, and E.~G. Learned{-}Miller.
\newblock Multi-view convolutional neural networks for 3d shape recognition.
\newblock In {\em Proc. ICCV}, 2015.

\bibitem{Maturana2015VoxNet}
D.~Maturana and S.~Scherer.
\newblock Voxnet: A 3d convolutional neural network for real-time object
  recognition.
\newblock In {\em Ieee/rsj International Conference on Intelligent Robots and
  Systems}, pages 922--928, 2015.

\bibitem{qi2017pointnet}
C.~R. Qi, H.~Su, K.~Mo, and L.~J. Guibas.
\newblock Pointnet: Deep learning on point sets for 3d classification and
  segmentation.
\newblock In {\em Proceedings of the IEEE conference on computer vision and
  pattern recognition}, pages 652--660, 2017.

\bibitem{qi2017pointnet++}
C.~R. Qi, L.~Yi, H.~Su, and L.~J. Guibas.
\newblock Pointnet++: Deep hierarchical feature learning on point sets in a
  metric space.
\newblock {\em Advances in neural information processing systems}, 30, 2017.

\bibitem{thomas2019KPConv}
H.~Thomas, C.~R. Qi, J.-E. Deschaud, B.~Marcotegui, F.~Goulette, and L.~J.
  Guibas.
\newblock Kpconv: Flexible and deformable convolution for point clouds.
\newblock In {\em Proceedings of the IEEE/CVF international conference on
  computer vision}, pages 6411--6420, 2019.

\bibitem{Riegler2017OctNet}
G.~Riegler, A.~O. Ulusoy, and A.~Geiger.
\newblock Octnet: Learning deep 3d representations at high resolutions.
\newblock In {\em Proceedings of the IEEE Conference on Computer Vision and
  Pattern Recognition}, 2017.

\bibitem{9652096}
S.~Huang, G.~Cai, Z.~Wang, Q.~Xia, and R.~Wang.
\newblock Ssa3d: Semantic segmentation assisted one-stage three-dimensional
  vehicle object detection.
\newblock {\em IEEE Transactions on Intelligent Transportation Systems},
  23(9):14764--14778, 2022.

\bibitem{9156722}
H.~Qi, C.~Feng, Z.~Cao, F.~Zhao, and Y.~Xiao.
\newblock P2b: Point-to-box network for 3d object tracking in point clouds.
\newblock In {\em 2020 IEEE/CVF Conference on Computer Vision and Pattern
  Recognition (CVPR)}, pages 6328--6337, 2020.

\bibitem{hu2019randla}
Q.~Hu, B.~Yang, L.~Xie, S.~Rosa, Y.~Guo, Z.~Wang, N.~Trigoni, and A.~Markham.
\newblock Randla-net: Efficient semantic segmentation of large-scale point
  clouds.
\newblock {\em Proceedings of the IEEE Conference on Computer Vision and
  Pattern Recognition}, 2020.

\bibitem{Loic18}
L.~Landrieu and M.~Simonovsky.
\newblock Large-scale point cloud semantic segmentation with superpoint graphs.
\newblock In {\em 2018 IEEE/CVF Conference on Computer Vision and Pattern
  Recognition}, pages 4558--4567, 2018.

\bibitem{chang2015shapenet}
A.~X. Chang, T.~Funkhouser, L.~Guibas, P.~Hanrahan, Q.~Huang, Z.~Li,
  S.~Savarese, M.~Savva, S.~Song, H.~Su, et~al.
\newblock Shapenet: An information-rich 3d model repository.
\newblock {\em arXiv preprint arXiv:1512.03012}, 2015.

\bibitem{uy2019revisiting}
M.~A. Uy, Q.-H. Pham, B.-S. Hua, T.~Nguyen, and S.-K. Yeung.
\newblock Revisiting point cloud classification: A new benchmark dataset and
  classification model on real-world data.
\newblock In {\em Proceedings of the IEEE/CVF international conference on
  computer vision}, pages 1588--1597, 2019.

\bibitem{6248074}
A.~Geiger, P.~Lenz, and R.~Urtasun.
\newblock Are we ready for autonomous driving? the kitti vision benchmark
  suite.
\newblock In {\em 2012 IEEE Conference on Computer Vision and Pattern
  Recognition}, pages 3354--3361, 2012.

\bibitem{9156412}
H.~Caesar, V.~Bankiti, A.~H. Lang, S.~Vora, V.~E. Liong, Q.~Xu, A.~Krishnan,
  Y.~Pan, G.~Baldan, and O.~Beijbom.
\newblock nuscenes: A multimodal dataset for autonomous driving.
\newblock In {\em 2020 IEEE/CVF Conference on Computer Vision and Pattern
  Recognition (CVPR)}, pages 11618--11628, 2020.

\bibitem{semantic3D2017}
T.~Hackel, N.~Savinov, L.~Ladicky, J.~Wegner, K.~Schindler, and M.~Pollefeys.
\newblock Semantic3d.net: A new large-scale point cloud classification
  benchmark.
\newblock {\em ISPRS Annals of Photogrammetry, Remote Sensing and Spatial
  Information Sciences}, IV-1/W1, 04 2017.

\bibitem{isprs}
F.~Rottensteiner, G.~Sohn, J.~Jung, M.~Gerke, C.~Baillard, S.~Bénitez, and
  U.~Breitkopf.
\newblock The isprs benchmark on urban object classification and 3d building
  reconstruction.
\newblock {\em ISPRS Annals of Photogrammetry, Remote Sensing and Spatial
  Information Sciences}, I-3, 07 2012.

\bibitem{hu2020towards}
Q.~Hu, B.~Yang, S.~Khalid, W.~Xiao, N.~Trigoni, and A.~Markham.
\newblock Towards semantic segmentation of urban-scale 3d point clouds: A
  dataset, benchmarks and challenges.
\newblock In {\em Proceedings of the IEEE/CVF Conference on Computer Vision and
  Pattern Recognition}, 2021.

\bibitem{Chen_2022_BMVC}
M.~Chen, Q.~Hu, Z.~Yu, H.~THOMAS, A.~Feng, Y.~Hou, K.~McCullough, F.~Ren, and
  L.~Soibelman.
\newblock Stpls3d: A large-scale synthetic and real aerial photogrammetry 3d
  point cloud dataset.
\newblock In {\em 33rd British Machine Vision Conference 2022, {BMVC} 2022,
  London, UK, November 21-24, 2022}. {BMVA} Press, 2022.

\bibitem{pmlr-v119-chen20j}
T.~Chen, S.~Kornblith, M.~Norouzi, and G.~Hinton.
\newblock A simple framework for contrastive learning of visual
  representations.
\newblock In H.~D. III and A.~Singh, editors, {\em Proceedings of the 37th
  International Conference on Machine Learning}, volume 119 of {\em Proceedings
  of Machine Learning Research}, pages 1597--1607. PMLR, 13--18 Jul 2020.

\bibitem{2018arXiv180703748V}
A.~{van den Oord}, Y.~{Li}, and O.~{Vinyals}.
\newblock {Representation Learning with Contrastive Predictive Coding}.
\newblock {\em arXiv e-prints}, page arXiv:1807.03748, July 2018.

\bibitem{DBLP:journals/corr/abs-1905-09272}
O.~J. H{\'{e}}naff, A.~Srinivas, J.~D. Fauw, A.~Razavi, C.~Doersch, S.~M.~A.
  Eslami, and A.~van~den Oord.
\newblock Data-efficient image recognition with contrastive predictive coding.
\newblock {\em CoRR}, abs/1905.09272, 2019.

\bibitem{sauder2019self}
J.~Sauder and B.~Sievers.
\newblock Self-supervised deep learning on point clouds by reconstructing
  space.
\newblock {\em Advances in Neural Information Processing Systems}, 32, 2019.

\bibitem{kim1999robust}
W.-Y. Kim and Y.-S. Kim.
\newblock Robust rotation angle estimator.
\newblock {\em IEEE Transactions on Pattern Analysis and Machine Intelligence},
  21(8):768--773, 1999.

\bibitem{xie2020pointcontrast}
S.~Xie, J.~Gu, D.~Guo, C.~R. Qi, L.~Guibas, and O.~Litany.
\newblock Pointcontrast: Unsupervised pre-training for 3d point cloud
  understanding.
\newblock In {\em Computer Vision--ECCV 2020: 16th European Conference,
  Glasgow, UK, August 23--28, 2020, Proceedings, Part III 16}, pages 574--591.
  Springer, 2020.

\bibitem{wang2021unsupervised}
H.~Wang, Q.~Liu, X.~Yue, J.~Lasenby, and M.~J. Kusner.
\newblock Unsupervised point cloud pre-training via occlusion completion.
\newblock In {\em Proceedings of the IEEE/CVF international conference on
  computer vision}, pages 9782--9792, 2021.

\bibitem{Liu2019ICLR}
Y.~Liu, S.~D'Aronco, and J.~Wegner.
\newblock Pc2wf: 3d wireframe reconstruction from raw point clouds.
\newblock In {\em International Conference on Learning Representations (ICLR)},
  03 2021.

\bibitem{wichmann2018roofn3d}
A.~Wichmann, A.~Agoub, and M.~Kada.
\newblock Roofn3d: Deep learning training data for 3d building reconstruction.
\newblock {\em International Archives of the Photogrammetry, Remote Sensing \&
  Spatial Information Sciences}, 42(2), 2018.

\bibitem{huang2022city3d}
J.~Huang, J.~Stoter, R.~Peters, and L.~Nan.
\newblock City3d: Large-scale building reconstruction from airborne lidar point
  clouds.
\newblock {\em Remote Sensing}, 14(9):2254, 2022.

\bibitem{lin2021capturing}
L.~Lin, Y.~Liu, Y.~Hu, X.~Yan, K.~Xie, and H.~Huang.
\newblock Capturing, reconstructing, and simulating: the urbanscene3d dataset.
\newblock {\em arXiv preprint arXiv:2107.04286}, 2021.

\bibitem{geiger2012we}
A.~Geiger, P.~Lenz, and R.~Urtasun.
\newblock Are we ready for autonomous driving? the kitti vision benchmark
  suite.
\newblock In {\em 2012 IEEE conference on computer vision and pattern
  recognition}, pages 3354--3361. IEEE, 2012.

\bibitem{li2022point2roof}
L.~Li, N.~Song, F.~Sun, X.~Liu, R.~Wang, J.~Yao, and S.~Cao.
\newblock Point2roof: End-to-end 3d building roof modeling from airborne lidar
  point clouds.
\newblock {\em ISPRS Journal of Photogrammetry and Remote Sensing}, 193:17--28,
  2022.

\bibitem{BMVC2019}
S.~M.~I. Zolanvari, S.~Ruano, A.~Rana, A.~Cummins, R.~E. da~Silva, M.~Rahbar,
  and A.~Smolic.
\newblock Dublincity: Annotated lidar point cloud and its applications.
\newblock In K.~Sidorov and Y.~Hicks, editors, {\em Proceedings of the British
  Machine Vision Conference (BMVC)}, pages 127.1--127.13. BMVA Press, September
  2019.

\bibitem{park2019deepsdf}
J.~J. Park, P.~Florence, J.~Straub, R.~Newcombe, and S.~Lovegrove.
\newblock Deepsdf: Learning continuous signed distance functions for shape
  representation.
\newblock In {\em Proceedings of the IEEE/CVF conference on computer vision and
  pattern recognition}, pages 165--174, 2019.

\bibitem{mescheder2019occupancy}
L.~Mescheder, M.~Oechsle, M.~Niemeyer, S.~Nowozin, and A.~Geiger.
\newblock Occupancy networks: Learning 3d reconstruction in function space.
\newblock In {\em Proceedings of the IEEE/CVF conference on computer vision and
  pattern recognition}, pages 4460--4470, 2019.

\bibitem{peng2020convolutional}
S.~Peng, M.~Niemeyer, L.~Mescheder, M.~Pollefeys, and A.~Geiger.
\newblock Convolutional occupancy networks.
\newblock In {\em European Conference on Computer Vision}, pages 523--540.
  Springer, 2020.

\bibitem{poullis2009automatic}
C.~Poullis and S.~You.
\newblock Automatic reconstruction of cities from remote sensor data.
\newblock In {\em 2009 IEEE conference on computer vision and pattern
  recognition}, pages 2775--2782. IEEE, 2009.

\bibitem{Nan2010}
L.~Nan, A.~Sharf, H.~Zhang, D.~Cohen-Or, and B.~Chen.
\newblock Smartboxes for interactive urban reconstruction.
\newblock {\em ACM Transactions on Graphics}, 29, 07 2010.

\bibitem{zhou20122}
Q.-Y. Zhou and U.~Neumann.
\newblock 2.5 d building modeling by discovering global regularities.
\newblock In {\em 2012 IEEE Conference on Computer Vision and Pattern
  Recognition}, pages 326--333. IEEE, 2012.

\bibitem{Vanegas2012}
C.~Vanegas, D.~Aliaga, and B.~Benes.
\newblock Automatic extraction of manhattan-world building masses from 3d laser
  range scans.
\newblock {\em IEEE transactions on visualization and computer graphics}, 18,
  01 2012.

\bibitem{henn2013model}
A.~Henn, G.~Gr{\"o}ger, V.~Stroh, and L.~Pl{\"u}mer.
\newblock Model driven reconstruction of roofs from sparse lidar point clouds.
\newblock {\em ISPRS Journal of photogrammetry and remote sensing}, 76:17--29,
  2013.

\bibitem{vosselman20013d}
G.~Vosselman, S.~Dijkman, et~al.
\newblock 3d building model reconstruction from point clouds and ground plans.
\newblock {\em International archives of photogrammetry remote sensing and
  spatial information sciences}, 34(3/W4):37--44, 2001.

\bibitem{chen2017topologically}
D.~Chen, R.~Wang, and J.~Peethambaran.
\newblock Topologically aware building rooftop reconstruction from airborne
  laser scanning point clouds.
\newblock {\em IEEE Transactions on Geoscience and Remote Sensing},
  55(12):7032--7052, 2017.

\bibitem{nan2017polyfit}
L.~Nan and P.~Wonka.
\newblock Polyfit: Polygonal surface reconstruction from point clouds.
\newblock In {\em Proceedings of the IEEE International Conference on Computer
  Vision}, pages 2353--2361, 2017.

\bibitem{PoullisPAMI2019}
C.~Poullis.
\newblock Large-scale urban reconstruction with tensor clustering and global
  boundary refinement.
\newblock {\em IEEE Transactions on Pattern Analysis and Machine Intelligence},
  42(5):1132--1145, 2020.

\bibitem{Minglei2016}
M.~Li, P.~Wonka, and L.~Nan.
\newblock Manhattan-world urban reconstruction from point clouds.
\newblock In {\em ECCV 2016}, volume 9908, pages 54--69, 10.

\bibitem{zhao2021point}
H.~Zhao, L.~Jiang, J.~Jia, P.~H. Torr, and V.~Koltun.
\newblock Point transformer.
\newblock In {\em Proceedings of the IEEE/CVF International Conference on
  Computer Vision}, pages 16259--16268, 2021.

\bibitem{zhou2020graph}
J.~Zhou, G.~Cui, S.~Hu, Z.~Zhang, C.~Yang, Z.~Liu, L.~Wang, C.~Li, and M.~Sun.
\newblock Graph neural networks: A review of methods and applications.
\newblock {\em AI open}, 1:57--81, 2020.

\bibitem{yu2022point}
X.~Yu, L.~Tang, Y.~Rao, T.~Huang, J.~Zhou, and J.~Lu.
\newblock Point-bert: Pre-training 3d point cloud transformers with masked
  point modeling.
\newblock In {\em Proceedings of the IEEE/CVF Conference on Computer Vision and
  Pattern Recognition}, pages 19313--19322, 2022.

\bibitem{pang2022masked}
Y.~Pang, W.~Wang, F.~E. Tay, W.~Liu, Y.~Tian, and L.~Yuan.
\newblock Masked autoencoders for point cloud self-supervised learning.
\newblock {\em In European Conference on Computer Vision}, 2022.

\bibitem{zhang2022point}
R.~Zhang, Z.~Guo, P.~Gao, R.~Fang, B.~Zhao, D.~Wang, Y.~Qiao, and H.~Li.
\newblock Point-m2ae: multi-scale masked autoencoders for hierarchical point
  cloud pre-training.
\newblock {\em arXiv preprint arXiv:2205.14401}, 2022.

\bibitem{zhou2022self}
J.~Zhou, X.~Wen, Y.-S. Liu, Y.~Fang, and Z.~Han.
\newblock Self-supervised point cloud representation learning with occlusion
  auto-encoder.
\newblock {\em arXiv preprint arXiv:2203.14084}, 2022.

\bibitem{phan2018dgcnn}
A.~V. Phan, M.~Le~Nguyen, Y.~L.~H. Nguyen, and L.~T. Bui.
\newblock Dgcnn: A convolutional neural network over large-scale labeled
  graphs.
\newblock {\em Neural Networks}, 108:533--543, 2018.

\bibitem{xu2021paconv}
M.~Xu, R.~Ding, H.~Zhao, and X.~Qi.
\newblock Paconv: Position adaptive convolution with dynamic kernel assembling
  on point clouds.
\newblock In {\em Proceedings of the IEEE/CVF Conference on Computer Vision and
  Pattern Recognition}, pages 3173--3182, 2021.

\bibitem{lai2022stratified}
X.~Lai, J.~Liu, L.~Jiang, L.~Wang, H.~Zhao, S.~Liu, X.~Qi, and J.~Jia.
\newblock Stratified transformer for 3d point cloud segmentation.
\newblock In {\em Proceedings of the IEEE/CVF Conference on Computer Vision and
  Pattern Recognition}, pages 8500--8509, 2022.

\bibitem{Liu2022FGNetAF}
K.~Liu, Z.~Gao, F.~Lin, and B.~Chen.
\newblock Fg-net: A fast and accurate framework for large-scale lidar point
  cloud understanding.
\newblock {\em IEEE Transactions on Cybernetics}, 53:553--564, 2022.

\bibitem{zhou20102}
Q.-Y. Zhou and U.~Neumann.
\newblock 2.5 d dual contouring: A robust approach to creating building models
  from aerial lidar point clouds.
\newblock In {\em European conference on computer vision}, pages 115--128.
  Springer, 2010.

\end{thebibliography}
}

\end{document}